\documentclass[runningheads]{llncs}
\usepackage[T1]{fontenc}
\usepackage{graphicx,verbatim}
\usepackage{amsfonts}

\usepackage{xcolor} 
\definecolor{darkgreen}{rgb}{0.0,0.5,0.3}
\usepackage[most]{tcolorbox}

\usepackage{booktabs}
\usepackage{tabularx}
\usepackage{multirow}
\usepackage{colortbl}
\usepackage{makecell}
\usepackage{siunitx}
\usepackage{hyperref}
\usepackage[symbol]{footmisc}

\usepackage{graphicx}

\definecolor{rankgold}{HTML}{FFD700}
\definecolor{ranksilver}{HTML}{C0C0C0}
\definecolor{rankbronze}{HTML}{CD7F32}

\newcommand{\first}[1]{\cellcolor{rankgold!25}\textbf{#1}}
\newcommand{\second}[1]{\cellcolor{ranksilver!30}{#1}}
\newcommand{\third}[1]{\cellcolor{rankbronze!18}{#1}}

\usepackage{enumitem}
\begin{document}
\title{EnTrust: Modeling Inter-Modal Conflict for Trustworthy Multimodal Medical Image Analysis}

\titlerunning{EnTrust}
%


\author{Dwarikanath Mahapatra$^{1,*}$,
Abhijit Das$^{2,*}$,
Behzad Bozorgtabar$^3$, \\
Zongyuan Ge$^4$,
Sudipta Roy$^5$,
Deepak Nayak$^6$,
Mauricio Reyes$^7$,  \\
Imran Razzak$^{2,8}$}
\authorrunning{Mahapatra et al.}
\institute{$^1$ Khalifa University, Abu Dhabi, UAE.
    $^2$ MBZUAI, Abu Dhabi, UAE. \\
    $^3$ Aarhus University, Denmark.
    $^4$ Monash University, Australia. \\
    $^5$ Jio University, Navi Mumbai, India.
    $^6$ NIT Jaipur, India.\\
    $^7$ University of Bern, Switzerland.
    $^8$ MedOS, Abu Dhabi, UAE.}
    
\maketitle
\let\thefootnote\relax\footnotetext{\textsuperscript{*}Equal contribution. Contact: \textit{dwarikanath.mahapatra@ku.ac.ae}}

\begin{abstract}
    Multimodal medical imaging fuses complementary anatomical and functional information, yet modalities frequently \emph{disagree} in pathologically heterogeneous regions. Current segmentation models handle this in one of two inadequate ways: deterministic fusion that averages away disagreement, or post-hoc uncertainty estimation decoupled from the fusion process that produces it. Both obscure the clinically critical question---\emph{why} is this prediction unreliable? We present \textbf{EnTrust}, a framework that treats inter-modal conflict as the primary source of predictive uncertainty. Our \textbf{EnFuse} module decomposes multimodal features into three disentangled components---shared anatomical consensus ($F_c$), modality-specific cues ($F_{u,m}$), and spatially localized conflict signals ($F_{cf}$)---with independence enforced via a cross-covariance objective. This structured decomposition conditions \textbf{SegDiff}, a diffusion-based generative segmentation model whose sampled hypotheses diverge specifically in regions of modal disagreement. \textbf{TrustMap} then translates this hypothesis divergence into calibrated, pixel-wise uncertainty using ensemble entropy, conflict-guided perturbation probing, and a learned calibration head---enabling clinicians to understand not only \emph{where} predictions are uncertain, but \emph{why}. Across four benchmarks spanning brain, cardiac, lesion, and oncology domains, EnTrust achieves state-of-the-art segmentation accuracy while reducing calibration error by 40\% compared to the strongest baseline. Notably, it outperforms 5$\times$ deep ensembles using a single model at roughly half the memory footprint. Code and checkpoints are available at \url{https://github.com/GenMI-Lab/EnTrust.git}. 

\end{abstract}

\section{Introduction}
\label{sec:introduction}

Clinical adoption of AI-assisted segmentation remains limited less by accuracy than by \emph{trust}.  A World Economic Forum report identifies low reliability as a major barrier to healthcare AI---particularly in resource-constrained settings~\cite{wef2024aihealthcare}---while an IBM survey finds that 43\% cite transparency concerns as the primary obstacle to clinical deployment~\cite{ibm2024aiadoption,human_centered_uq_2023}.  Radiologists routinely fuse complementary modalities (MRI sequences, CT, PET) to reach diagnostic judgments, yet these modalities frequently \emph{disagree} in heterogeneous pathologies~\cite{deep_ensembles,mc_dropout,das2024confidence}.  A trustworthy system must surface this disagreement---not suppress.

    \begin{figure*}[t]
        \centering
        \begin{minipage}[t]{0.49\textwidth}
            \centering
            \includegraphics[width=\textwidth]{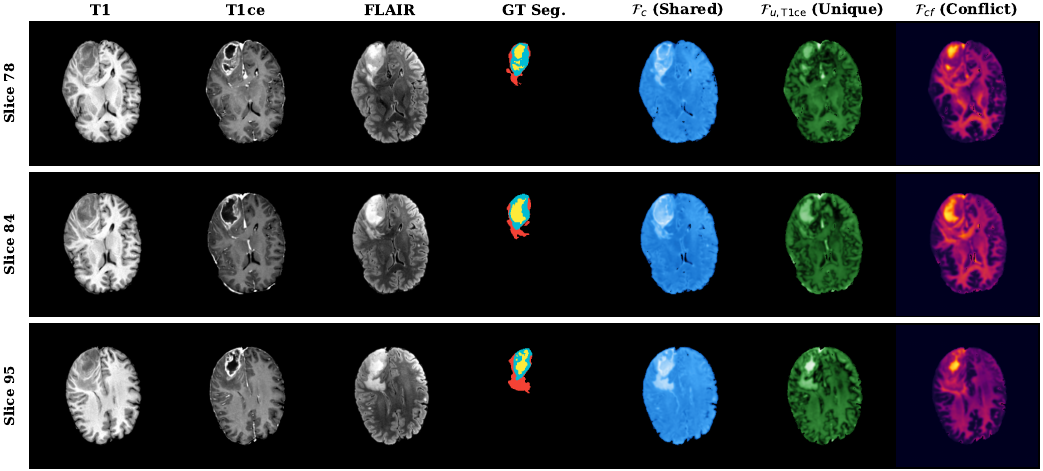}
            \tiny\textbf{(a) BraTS 2021 (Patient 00495)}\\
        \end{minipage}\hfill
        \begin{minipage}[t]{0.49\textwidth}
            \centering
            \includegraphics[width=\textwidth]{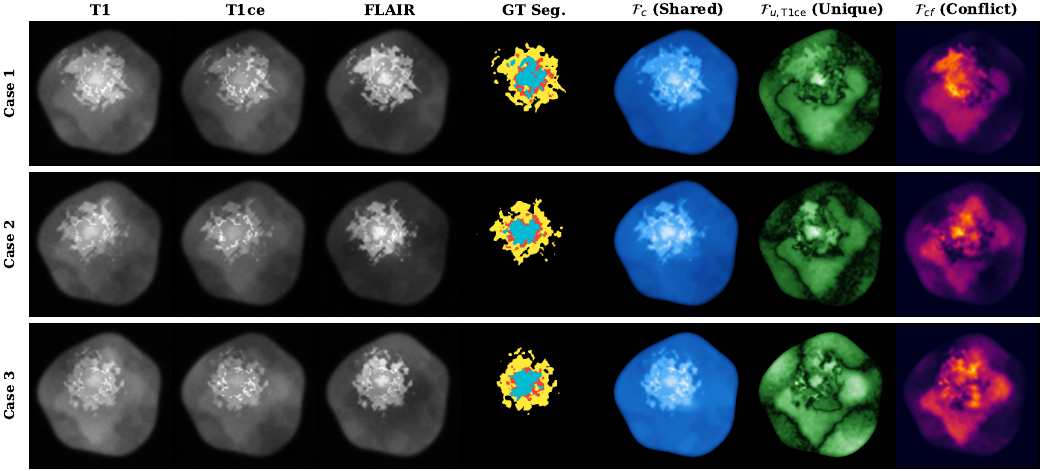}
            \tiny\textbf{(b) Synthetic Phantom}\\
        \end{minipage}
        \caption{EnFuse disentanglement on \textbf{(a)}~real BraTS 2021 MRI and \textbf{(b)}~synthetic phantoms. $\mathcal{F}_c$ captures modality-invariant anatomy, $\mathcal{F}_{u,\mathrm{T1ce}}$ isolates T1ce-specific contrast enhancement, and $\mathcal{F}_{cf}$ localizes inter-modal disagreement at tumor boundaries.}
        \label{fig:disentanglement}
    \end{figure*}

\noindent\textbf{Two inadequate paradigms.}\quad Existing approaches fall into two categories, both insufficient.  \emph{Deterministic fusion} methods---concatenation, attention-based gating---implicitly assume cross-modal consistency, averaging away the disagreement signals that carry the most clinical information~\cite{milletari2016v}.  \emph{Post-hoc uncertainty} methods---MC dropout~\cite{mc_dropout}, deep ensembles~\cite{deep_ensembles}, evidential networks~\cite{sensoy2018deep}---estimate uncertainty independently of fusion, so their estimates reflect model noise rather than the inter-modal conflict that drives it.  Neither paradigm tells the clinician \emph{which} modalities disagree or \emph{why} the prediction is unreliable. To address this research gap, We introduce \textbf{EnTrust} (Fig.~\ref{fig:architecture_diagram}, Section~\ref{sec:method}), built on the principle that inter-modal disagreement is a first-class modeling objective. It comprises three tightly coupled stages:

\begin{enumerate}
\item \textbf{EnFuse} disentangles multimodal features into shared anatomy, modality-specific cues, and spatially localized conflict---explicitly separating \emph{where modalities agree} from \emph{where and how they disagree} (Fig.~\ref{fig:disentanglement}).
\item \textbf{SegDiff} uses this structured decomposition to condition a diffusion-based segmentation model, so that sampled hypotheses diverge precisely in conflict regions rather than uniformly across the volume.
\item \textbf{TrustMap} converts hypothesis divergence into calibrated, pixel-wise uncertainty maps that tell clinicians not just \emph{where} predictions are uncertain but \emph{why}---grounded in the conflict signal from EnFuse.
\end{enumerate}

\section{Method}
\label{sec:method}

\begin{figure}[h]
        \centering
        \includegraphics[width=0.8\textwidth]{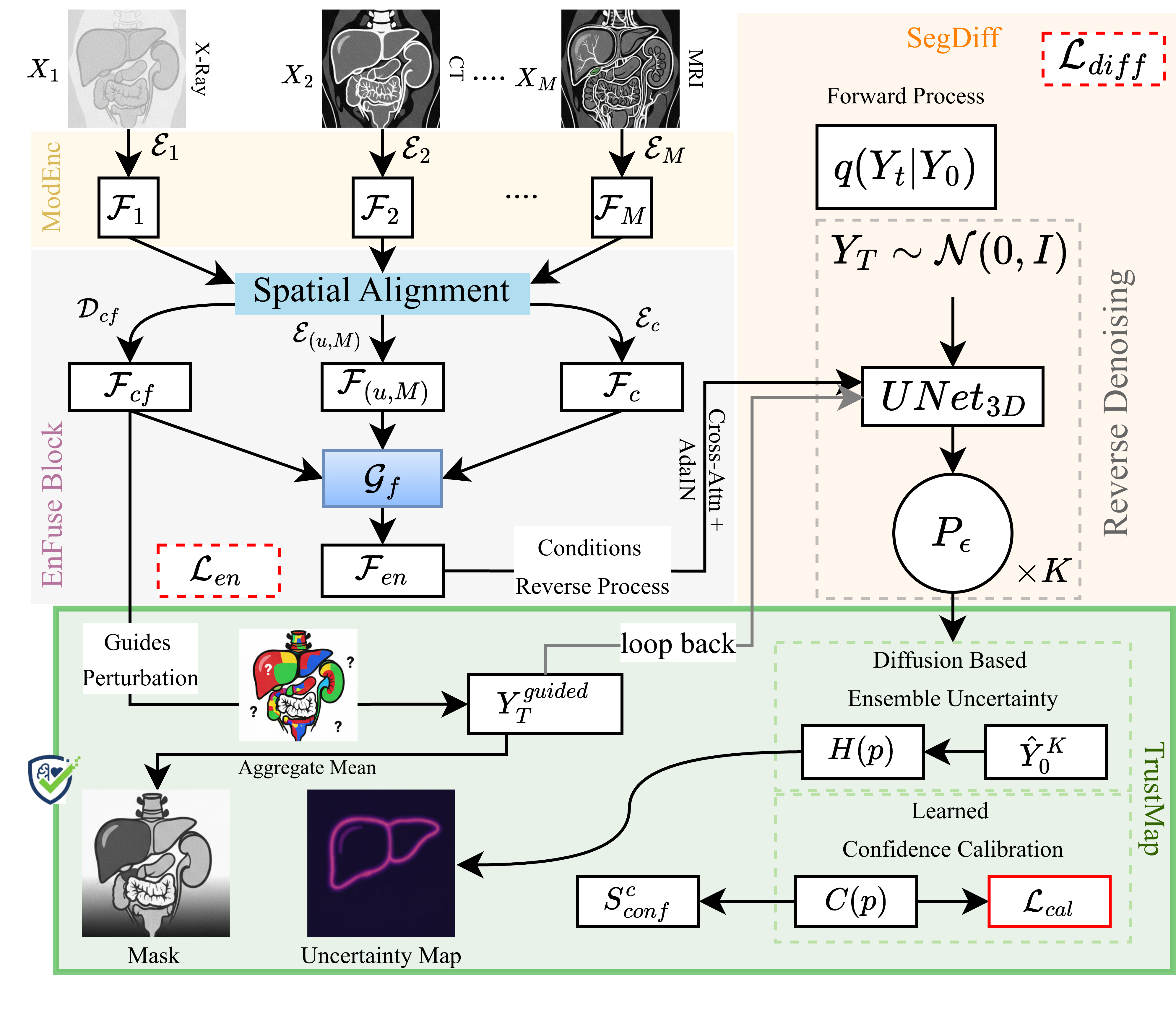}
        \caption{\textbf{EnTrust architecture.} ModEnc extracts per-modality features, which EnFuse disentangles into shared ($\mathcal{F}_c$), unique ($\mathcal{F}_{u,m}$), and conflict ($\mathcal{F}_{cf}$) components. The gated fusion $\mathcal{F}_{en}$ conditions SegDiff for probabilistic mask generation. TrustMap derives pixel-wise uncertainty via ensemble entropy $H(\mathbf{p})$, ProbeStep $\hat{Y}_0^K$, and CalibHead $C(\mathbf{p})$.}
        \label{fig:architecture_diagram}
    \end{figure}

EnTrust comprises four stages (Fig.~\ref{fig:architecture_diagram}): modality encoding (ModEnc), disentangled fusion (EnFuse), conditional diffusion segmentation (SegDiff), and uncertainty quantification (TrustMap), trained end-to-end with a composite loss.

\begin{tcolorbox}[colback=green!3,colframe=darkgreen!40,boxsep=2pt,left=3pt,right=3pt,top=3pt,bottom=3pt]
\small
\textbf{Contributions:}
\begin{itemize}[leftmargin=*,itemsep=1pt,topsep=2pt]
    \item \textbf{EnFuse}---to our knowledge, the first fusion module that explicitly decomposes multimodal features into consensus, unique, and conflict components with enforced statistical independence, making inter-modal disagreement a direct input to segmentation (\S\ref{subsec:enfuse}).
    \item \textbf{SegDiff + TrustMap}---the first pipeline to condition diffusion-based segmentation on structured conflict representations and distill hypothesis divergence into \emph{explainable}, calibrated pixel-wise uncertainty (\S\ref{subsec:segdiff}--\ref{subsec:trustmap}).
    \item \textbf{Practical impact}---SOTA accuracy and 40\% calibration improvement across four benchmarks, outperforming $5{\times}$ ensembles as a single model at ${\sim}$half the memory, with an encoder-agnostic design supporting drop-in backbone replacement (Table~\ref{tab:ablation_study}).
\end{itemize}
\end{tcolorbox}

\subsection{ModEnc: Modality Encoding}
\label{subsec:modenc}

Each modality $X_m$ is processed by a dedicated Swin Transformer encoder $\mathcal{E}_m$~\cite{SwinTrans}, producing feature maps $\mathcal{F}_m$ that are spatially aligned before fusion. Encoders are independent by design---any pretrained encoder can be substituted without modifying downstream modules (Table~\ref{tab:ablation_study}, blue block).

\subsection{EnFuse: Uncertainty-Aware Disentangled Fusion}
\label{subsec:enfuse}

Given feature representations $\{\mathcal{F}_1, \dots, \mathcal{F}_M\}$ from ModEnc, EnFuse decomposes them into three structurally distinct components without assuming cross-modal consistency. \textbf{EnFuse-Common} ($\mathcal{E}_c$) extracts modality-invariant features, producing the shared anatomical consensus $\mathcal{F}_c$. \textbf{EnFuse-Unique} ($\mathcal{E}_{u,m}$) isolates per-modality features not observable in the shared space, producing $\mathcal{F}_{u,m}$. \textbf{EnFuse-Conflict} ($\mathcal{D}_{cf}$)---our key contribution---identifies spatial regions where modalities provide contradictory evidence, outputting a conflict map $\mathcal{F}_{cf}$ that serves as an explicit uncertainty signal for downstream generation. The three components are integrated through \textbf{EnFuse-Gate} ($\mathcal{G}_f$) into a unified conditioning representation $\mathcal{F}_{en}$, where the inclusion of $\mathcal{F}_{cf}$ ensures conflict-driven uncertainty directly influences hypothesis generation. To enforce disentanglement, we minimize pairwise cross-covariance between all component pairs:
    \begin{equation}
    \begin{split}
    \mathcal{L}_{\text{en}} = & \sum_{m=1}^{M} \| \text{Cov}(\text{Vec}(\mathcal{F}_{c}),\; \text{Vec}(\mathcal{F}_{u,m})) \|_F^2 \\
    & + \sum_{m=1}^{M} \| \text{Cov}(\text{Vec}(\mathcal{F}_{u,m}),\; \text{Vec}(\mathcal{F}_{cf})) \|_F^2 + \| \text{Cov}(\text{Vec}(\mathcal{F}_{c}),\; \text{Vec}(\mathcal{F}_{cf})) \|_F^2
    \end{split}
    \label{eq:enfuse_loss}
    \end{equation}
Figure~\ref{fig:disentanglement} validates this decomposition on both real BraTS data and synthetic phantoms with known tissue boundaries.
\textbf{Implementation details}. All modality features are trilinearly resampled to the Swin stage-3 grid $(B\times C\times16^3)$ before fusion. $E_c$, $E_{u,m}$, and $D_{cf}$ are lightweight three-block ConvNeXt heads: $E_c$ receives all aligned modalities, $E_{u,m}$ is modality-indexed, and $D_{cf}$ receives aligned pairwise disagreement features. The EnFuse-Gate computes voxel-wise weights $w_i=\mathrm{softmax}(\mathrm{Conv}_{1\times1\times1}([F_c,\{F_{u,m}\},F_{cf}]))_i$ and fuses components as $F_{en}=\sum_i w_iF_i$. For Eq. 1, each component is average-pooled, projected to $C/4$ channels, mean-centered, reshaped to $N\times C'$, and cross-covariance is computed as $A^\top B/(N-1)$ with the Frobenius penalty normalized by $C'^2$.

\subsection{SegDiff: Conditional Diffusion Segmentation}
\label{subsec:segdiff}

SegDiff formulates segmentation as a conditional generative process using a standard DDPM~\cite{ho2020denoising} with a 3D U-Net $\epsilon_\theta$ as the denoiser. The forward process adds Gaussian noise to the ground-truth mask $Y_0$ over $T$ timesteps: $q(Y_t | Y_0) = \mathcal{N}(Y_t;\; \sqrt{\bar{\alpha}_t}\, Y_0,\; (1 - \bar{\alpha}_t)I)$. The reverse process denoises $Y_T \sim \mathcal{N}(0, I)$ back to $Y_0$, conditioned on $\mathcal{F}_{en}$ via cross-attention and AdaIN:
    \begin{equation}
    \mathcal{L}_{\text{diff}} = \mathbb{E}_{t, Y_0, \epsilon} \Big[\big\| \epsilon - \epsilon_\theta(\sqrt{\bar{\alpha}_t}Y_0 + \sqrt{1-\bar{\alpha}_t}\,\epsilon,\; t,\; \mathcal{F}_{en}) \big\|^2 \Big]
    \label{eq:segdiff_loss}
    \end{equation}
What distinguishes SegDiff from prior diffusion segmentation~\cite{medsegdiff_2024} is its conditioning: $\mathcal{F}_{en}$ carries structured conflict information, so the $K$ hypotheses $\{\hat{Y}_0^{(1)}, \dots, \hat{Y}_0^{(K)}\}$ sampled at inference diverge specifically where modalities disagree---encoding clinically meaningful uncertainty rather than random variation.

\subsection{TrustMap: Uncertainty Quantification}
\label{subsec:trustmap}

TrustMap derives pixel-wise uncertainty through three complementary mechanisms \textemdash Ensemble Entropy, ProbeStep and Calibration Head.


\noindent\textbf{Ensemble Entropy.}. For voxel $v$, class probabilities are averaged over $K$ reverse
diffusion samples, $\hat p_c(v)=K^{-1}\sum_{k=1}^K
\mathrm{softmax}(\hat Y^{(k)}_0)_c(v)$. Predictive entropy
$H(v)=-\sum_c \hat p_c(v)\log\hat p_c(v)$ measures total predictive uncertainty.
We additionally compute sample dispersion $\sigma_K(v)$; CalibHead receives
$[H(v),\sigma_K(v), \linebreak \max_c\hat p_c(v),F_{cf}(v)]$ to calibrate confidence.

\noindent\textbf{ProbeStep.} Entropy may underestimate uncertainty where the model is confidently wrong. ProbeStep perturbs the initial noise using the conflict map to stress-test such regions:
\begin{equation}
Y_T^{\text{guided}} = Y_T + \gamma \cdot \text{perturbation\_map}(\mathcal{F}_{cf})
\label{eq:probestep}
\end{equation}
The perturbed $Y_T^{\text{guided}}$ produces additional samples that probe sensitivity at conflict boundaries, applied solely at inference with no retraining.

\noindent\textbf{CalibHead.} An auxiliary head predicts pixel-wise confidence $C(\mathbf{p})$, trained with:
\begin{equation}
\mathcal{L}_{\text{cal}} = \mathbb{E}_{\mathbf{p}} \left[ \left| C(\mathbf{p}) - \mathbb{I}(\hat{Y}_0(\mathbf{p}) = Y_{GT}(\mathbf{p})) \right| \right]
\label{eq:calibhead_loss}
\end{equation}
This ensures predicted confidence aligns with actual correctness---when EnTrust reports 80\% confidence, the prediction is correct $\sim$80\% of the time.

\section{Experiments}
\label{sec:experiments}

\subsection{Benchmarks and Baselines}
\label{subsec:datasets}

We evaluate on four multimodal benchmarks (Table~\ref{tab:datasets}) against five methods spanning the major uncertainty paradigms: 3D U-Net~\cite{milletari2016v} (deterministic fusion), Probabilistic U-Net~\cite{kohl2018probabilistic} (conditional VAE), Deep Ensemble~\cite{deep_ensembles} ($5\times$ models), DEviS~\cite{zou2024reliablemedicalimagesegmentation} (evidential), and MedSegDiff~\cite{medsegdiff_2024} (diffusion-based). We report DSC, HD95, ECE, AUROC-Error, and NLL; UQ metrics are omitted for U-Net.

\begin{table*}[t]
    \centering
    \caption{\textbf{Evaluation benchmarks}. Splits fixed across all methods, released with code.}
    \label{tab:datasets}
    \renewcommand{\arraystretch}{1.15}
    \scriptsize
    \resizebox{\textwidth}{!}{%
    \begin{tabular}{@{} l l l r l @{}}
        \toprule
        \textbf{Dataset} & \textbf{Modalities} & \textbf{Segmentation Task} & \textbf{Train\,/\,Val\,/\,Test} & \textbf{Split Strategy} \\
        \midrule
        \rowcolor{gray!4}
        BraTS 2021~\cite{brats_dataset}                       & T1, T1ce, T2, FLAIR & Glioma (brain)          & 1{,}000\,/\,125\,/\,126 & Stratified by tumor grade \\
        MS-CMRSeg~\cite{zhuang2018multivariate}               & bSSFP, LGE          & Cardiac structures      & 25\,/\,5\,/\,15          & Official challenge split \\
        \rowcolor{gray!4}
        ATLAS~\cite{fedorov2019atlas}                         & CT, FLAIR MRI       & Brain lesion            & 524\,/\,65\,/\,66        & Stratified by lesion vol. \\
        H\&N CT-PET~\cite{headneckdataset}                            & CT, PET             & Head \& neck tumor      & 156\,/\,34\,/\,34        & HECKTOR challenge split \\
        \bottomrule
        \addlinespace[2pt]
        \multicolumn{5}{@{}l}{\textcolor{gray}{\scriptsize\textit{Preprocessing: isotropic resampling, per-modality z-score normalization, uniform cropping to $128^3$ voxels.}}}
    \end{tabular}%
    }
\end{table*}

\subsection{Implementation Details}
\label{subsec:implementation}

EnTrust is implemented in PyTorch and trained end-to-end on $4\times$ NVIDIA A100 GPUs (64\,GB) for 300 epochs with batch size 2. The diffusion backbone uses $T{=}1000$ forward timesteps with a cosine noise schedule; at inference, $K{=}5$ stochastic reverse samples are drawn per input (sensitivity to $K$ is analyzed in the supplementary, Table~S1). Optimization uses AdamW (lr from $1\times10^{-4}$ to $1\times10^{-6}$, cosine annealing). The composite training loss is:
\begin{equation}
    \mathcal{L}_{\text{total}} = \mathcal{L}_{\text{diff}} + \lambda_{\text{en}}\,\mathcal{L}_{\text{en}} + \lambda_{\text{cal}}\,\mathcal{L}_{\text{cal}}
    \label{eq:total_loss}
\end{equation}
where $\mathcal{L}_{\text{diff}}$ drives segmentation (Eq.~\ref{eq:segdiff_loss}), $\mathcal{L}_{\text{en}}$ enforces disentanglement (Eq.~\ref{eq:enfuse_loss}), and $\mathcal{L}_{\text{cal}}$ ensures calibration (Eq.~\ref{eq:calibhead_loss}). Loss weights $\lambda_{\text{en}}{=}0.1$ and $\lambda_{\text{cal}}{=}0.5$ are selected via grid search on the BraTS 2021 validation set (Table~\ref{tab:lambda_sensitivity}) and held fixed across all four datasets without per-dataset tuning. All results are averaged over 3 independent runs with seeds $\{42, 123, 256\}$ using identical splits. Code, checkpoints, split files, and configurations will be released upon acceptance.

\begin{table*}[t]
    \centering
    \tiny
    \caption{Sensitivity of loss weights on BraTS 2021 validation set. Left: varying $\lambda_{\text{en}}$ ($\lambda_{\text{cal}}{=}0.5$). Right: varying $\lambda_{\text{cal}}$ ($\lambda_{\text{en}}{=}0.1$). Selected configuration in \colorbox{green!15}{green}.}
    \label{tab:lambda_sensitivity}
    \resizebox{\textwidth}{!}{%
    \begin{tabular}{@{} l cccc c cccc @{}}
        \toprule
        & \multicolumn{4}{c}{$\lambda_{\text{cal}} = 0.5$ (varying $\lambda_{\text{en}}$)} & & \multicolumn{4}{c}{$\lambda_{\text{en}} = 0.1$ (varying $\lambda_{\text{cal}}$)} \\
        \cmidrule(lr){2-5} \cmidrule(l){7-10}
        \textbf{Metric} & $\lambda_{\text{en}}{=}0.01$ & $\lambda_{\text{en}}{=}0.05$ & \cellcolor{green!15} $\lambda_{\text{en}}{=}0.10$ & $\lambda_{\text{en}}{=}0.50$ & & $\lambda_{\text{cal}}{=}0.1$ & \cellcolor{green!15} $\lambda_{\text{cal}}{=}0.5$ & $\lambda_{\text{cal}}{=}1.0$ & $\lambda_{\text{cal}}{=}2.0$ \\
        \midrule
        DSC ($\uparrow$)   & 0.878 & 0.885 & \cellcolor{green!15} \textbf{0.892} & 0.880 & & 0.891 & \cellcolor{green!15} \textbf{0.892} & 0.888 & 0.875 \\
        ECE ($\downarrow$)   & 0.072 & 0.068 & \cellcolor{green!15} \textbf{0.061} & 0.065 & & 0.095 & \cellcolor{green!15} \textbf{0.061} & 0.058 & 0.055 \\
        AUROC ($\uparrow$) & 0.875 & 0.882 & \cellcolor{green!15} \textbf{0.890} & 0.878 & & 0.872 & \cellcolor{green!15} \textbf{0.890} & 0.888 & 0.880 \\
        \bottomrule
    \end{tabular}%
    }
\end{table*}

\section{Results and Discussion}
\label{subsec:results}

    \begin{table*}[t]
        \centering
        \tiny
        \caption{Segmentation accuracy and uncertainty quantification across four multimodal benchmarks. Cell shading: \colorbox{rankgold!25}{\textbf{1st}}, \colorbox{ranksilver!30}{2nd}, \colorbox{rankbronze!18}{3rd}. UQ metrics omitted for U-Net. All values: mean $\pm$ std over 3 runs.}
        \label{tab:overall_performance}
        \resizebox{\textwidth}{!}{%
        \begin{tabular}{@{} l l c c c c c @{}}
            \toprule
            \textbf{Method} & \textbf{Dataset} & \textbf{DSC} ($\uparrow$) & \textbf{HD95} ($\downarrow$) & \textbf{ECE} ($\downarrow$) & \textbf{AUROC-Err} ($\uparrow$) & \textbf{NLL} ($\downarrow$) \\
            \midrule
    
            \multirow{6}{*}{\rotatebox{90}{\textbf{BraTS}}}
            & U-Net (Early Fusion)   & 0.782 $\pm$ 0.051 & 15.2 $\pm$ 3.1 & --    & --    & --    \\
            & Probabilistic U-Net    & 0.821 $\pm$ 0.042 & 12.8 $\pm$ 2.5 & 0.183 & 0.751 & 0.320 \\
            & Deep Ensemble (5$\times$) & \third{0.850 $\pm$ 0.031} & \third{11.5 $\pm$ 2.0} & \third{0.121} & \second{0.812} & \third{0.250} \\
            & DEviS                  & 0.835 $\pm$ 0.038 & 12.0 $\pm$ 2.2 & 0.152 & 0.785 & 0.295 \\
            & MedSegDiff             & \second{0.865 $\pm$ 0.028} & \second{10.8 $\pm$ 1.8} & \second{0.101} & \third{0.830} & \second{0.235} \\
            & \textbf{EnTrust (Ours)} & \first{0.892 $\pm$ 0.020} & \first{9.8 $\pm$ 1.5} & \first{0.061} & \first{0.890} & \first{0.180} \\
    
            \midrule
    
            \multirow{6}{*}{\rotatebox{90}{\textbf{CMRSeg}}}
            & U-Net (Early Fusion)   & 0.815 $\pm$ 0.045 & 12.8 $\pm$ 2.9 & --    & --    & --    \\
            & Probabilistic U-Net    & 0.852 $\pm$ 0.038 & 10.5 $\pm$ 2.1 & 0.165 & 0.780 & 0.285 \\
            & Deep Ensemble (5$\times$) & \third{0.875 $\pm$ 0.028} & \third{9.8 $\pm$ 1.8} & \second{0.105} & \second{0.835} & \third{0.220} \\
            & DEviS                  & 0.860 $\pm$ 0.032 & 10.2 $\pm$ 2.0 & 0.140 & 0.810 & 0.260 \\
            & MedSegDiff             & \second{0.880 $\pm$ 0.025} & \second{9.5 $\pm$ 1.6} & \third{0.090} & \third{0.850} & \second{0.205} \\
            & \textbf{EnTrust (Ours)} & \first{0.910 $\pm$ 0.018} & \first{8.5 $\pm$ 1.2} & \first{0.055} & \first{0.905} & \first{0.150} \\
    
            \midrule
    
            \multirow{6}{*}{\rotatebox{90}{\textbf{ATLAS}}}
            & U-Net (Early Fusion)   & 0.755 $\pm$ 0.062 & 18.1 $\pm$ 4.2 & --    & --    & --    \\
            & Probabilistic U-Net    & 0.790 $\pm$ 0.055 & 14.5 $\pm$ 3.5 & 0.198 & 0.735 & 0.355 \\
            & Deep Ensemble (5$\times$) & \third{0.825 $\pm$ 0.045} & \second{12.0 $\pm$ 2.8} & \second{0.140} & \third{0.790} & \third{0.290} \\
            & DEviS                  & 0.805 $\pm$ 0.050 & 13.0 $\pm$ 3.0 & 0.165 & 0.770 & 0.320 \\
            & MedSegDiff             & \second{0.835 $\pm$ 0.040} & \third{11.5 $\pm$ 2.5} & \third{0.120} & \second{0.805} & \second{0.275} \\
            & \textbf{EnTrust (Ours)} & \first{0.875 $\pm$ 0.030} & \first{10.0 $\pm$ 2.0} & \first{0.075} & \first{0.865} & \first{0.210} \\
    
            \midrule
    
            \multirow{6}{*}{\rotatebox{90}{\textbf{H\&N}}}
            & U-Net (Early Fusion)   & 0.778 $\pm$ 0.055 & 16.5 $\pm$ 3.8 & --    & --    & --    \\
            & Probabilistic U-Net    & 0.805 $\pm$ 0.048 & 13.8 $\pm$ 3.1 & 0.175 & 0.762 & 0.310 \\
            & Deep Ensemble (5$\times$) & \third{0.838 $\pm$ 0.035} & \third{11.2 $\pm$ 2.5} & \third{0.130} & \third{0.805} & \third{0.270} \\
            & DEviS                  & 0.818 $\pm$ 0.040 & 12.5 $\pm$ 2.8 & 0.155 & 0.790 & 0.290 \\
            & MedSegDiff             & \second{0.848 $\pm$ 0.030} & \second{10.8 $\pm$ 2.2} & \second{0.110} & \second{0.820} & \second{0.250} \\
            & \textbf{EnTrust (Ours)} & \first{0.885 $\pm$ 0.025} & \first{9.5 $\pm$ 1.8} & \first{0.070} & \first{0.870} & \first{0.190} \\
    
            \bottomrule
        \end{tabular}
        }
    \end{table*}

        \begin{figure}[t]
            \centering
            \includegraphics[width=0.8\linewidth]{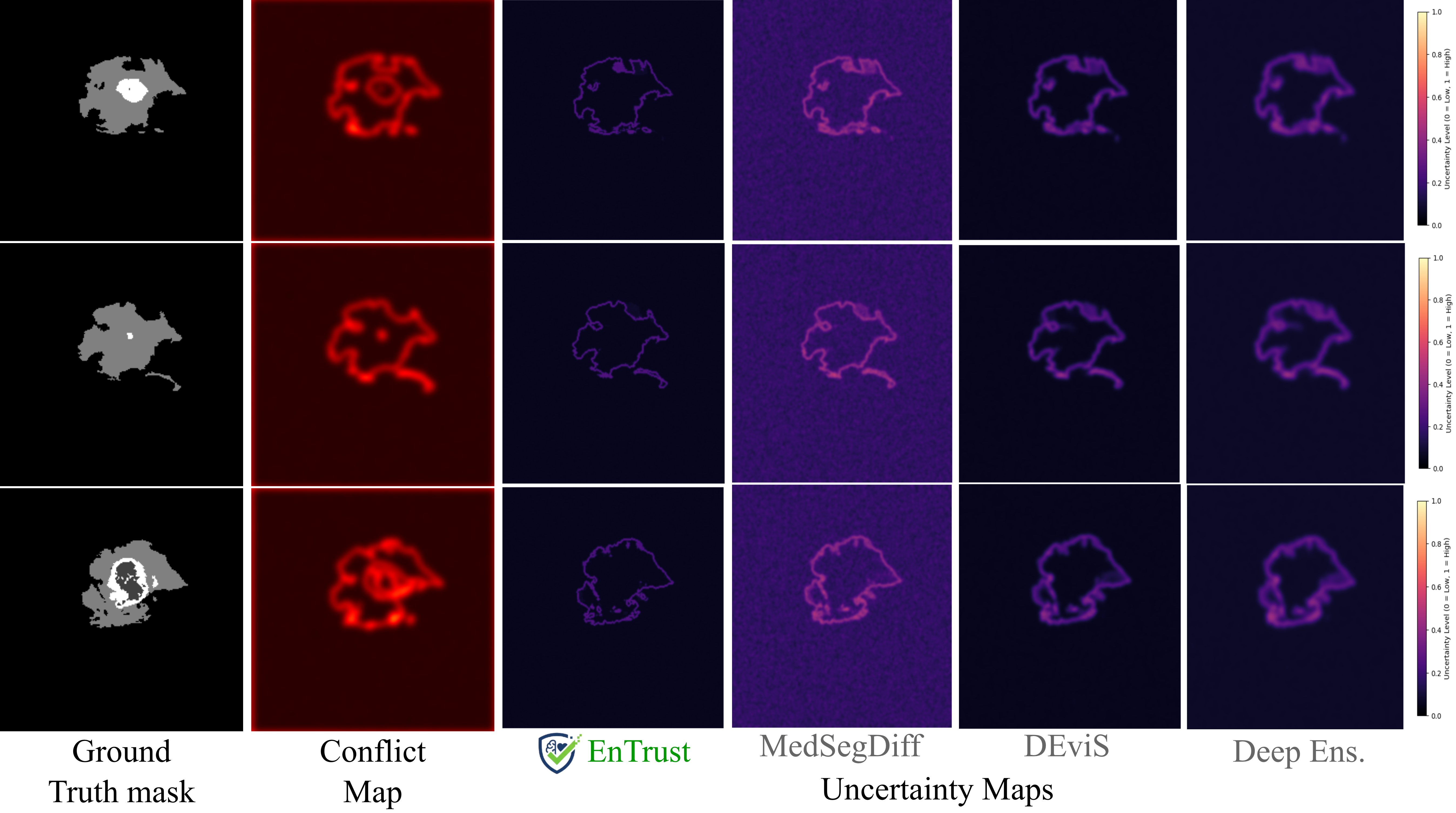}
            \caption{\textbf{Qualitative comparison on BraTS 2021.} EnTrust produces boundary-localized uncertainty aligned with $\mathcal{F}_{cf}$; baselines exhibit progressively diffuse estimates.}
            \label{fig:qualitative_results}
        \end{figure}


    \subsection{State-of-the-Art Performance}
        
\begin{figure}[t]
            \centering
            \includegraphics[width=1\linewidth]{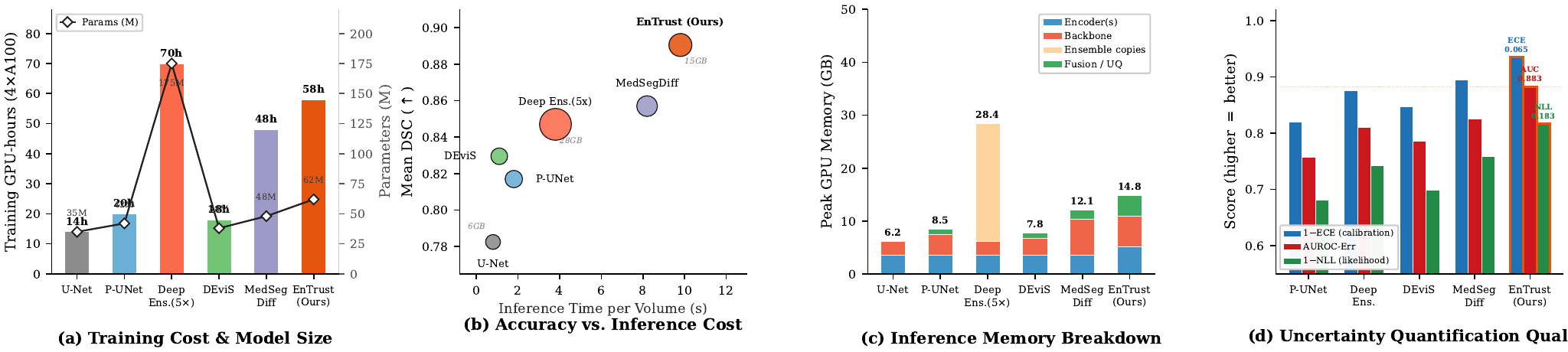}
            \caption{\textbf{Computational analysis.} \textbf{(a)}~Training cost and model size. \textbf{(b)}~Accuracy vs.\ inference cost (bubble $=$ peak GPU memory). \textbf{(c)}~Inference memory breakdown. \textbf{(d)}~UQ quality comparison.}
            \label{fig:computational_analysis}
        \end{figure}

        Table~\ref{tab:overall_performance} shows that EnTrust obtains the best mean performance on all reported
metrics across the four datasets. Paired Wilcoxon tests on case-wise DSC, averaged
over three seeds, show significant gains over MedSegDiff and Deep Ensemble on
BraTS, ATLAS, and H$\&$N.
        Two controlled comparisons are particularly informative. Against MedSegDiff, which shares the same 3D U-Net diffusion backbone, EnTrust consistently improves both accuracy and calibration across all benchmarks; since the denoiser and noise schedule are identical, these gains isolate the contribution of EnFuse and TrustMap. Against Deep Ensemble, EnTrust achieves superior performance with a single model versus five independently trained networks, at roughly half the memory and 17\% less training time (Fig.~\ref{fig:computational_analysis}). Figure~\ref{fig:qualitative_results} corroborates this quantitatively: EnTrust uncertainty maps are tightly boundary-localized and align with the conflict map $\mathcal{F}_{cf}$, whereas baseline methods produce progressively diffuse estimates lacking spatial precision for clinical use.

        \begin{tcolorbox}[colback=green!2,colframe=darkgreen!50,boxsep=1pt,left=3pt,right=3pt,top=2pt,bottom=2pt,title={\textbf{Discussion: Why EnTrust Matters}},fonttitle=\bfseries\scriptsize]
        \small
            \textbf{$\odot$ Uncertainty at the right level.} Post-hoc methods quantify \textit{model} uncertainty---how much predictions vary. EnTrust quantifies \textit{data} uncertainty---where input evidence is contradictory. A radiologist seeing T1ce--FLAIR conflict at a tumor margin can order additional sequences; knowing the network is ``not confident'' offers no actionable recourse.
            
            \textbf{$\odot$ One model, three clinical answers.} $\mathcal{F}_c$ shows where modalities agree, $\mathcal{F}_{u,m}$ reveals what each uniquely contributes, and $\mathcal{F}_{cf}$ pinpoints where they conflict. No prior method provides this structured decomposition from a single forward pass.

            \textbf{$\odot$ Efficiency without compromise.} EnTrust matches or exceeds $5\times$ ensemble accuracy at $1\times$ training cost, $3\times$ fewer parameters, and $\sim$half the inference memory (Fig.~\ref{fig:computational_analysis}).
        \end{tcolorbox}

    \subsection{Ablation Study}
        Table~\ref{tab:ablation_study} isolates each component via systematic removal, with $\overline{\Delta}$ rows quantifying mean degradation. EnFuse removal causes the largest collapse across every metric, confirming disentangled fusion as foundational---without it, the model reverts to implicit fusion that conflates consensus and contradiction. Removing only $\mathcal{D}_{cf}$ degrades both accuracy and calibration, showing that conflict modeling improves segmentation quality, not just uncertainty. CalibHead removal leaves DSC/HD95 unchanged but degrades ECE by 58.2\%---a clean separation confirming that accuracy and calibration are decoupled objectives. ProbeStep removal minimally affects DSC but consistently worsens ECE, validating its targeted role at conflict boundaries. The encoder modularity block (blue rows) demonstrates that swapping Swin-T for ResNet-50 or ViT-B/16 without modifying EnFuse or TrustMap yields competitive results, confirming encoder-agnostic design.
        \textbf{Limitations}. TrustMap improves spatial localization of uncertainty, but clinical
utility beyond expected boundary uncertainty should be validated in reader studies.

    \begin{table*}[t]
    \centering
    \caption{Ablation study on EnTrust components and encoder modularity. \colorbox{green!8}{Green}: full model. \colorbox{red!8}{Red}: worst degradation. \colorbox{blue!6}{Blue}: encoder variants. $\overline{\Delta}$:\% change from EnTrust.}
    \label{tab:ablation_study}
    \setlength{\tabcolsep}{3.5pt}
    \renewcommand{\arraystretch}{1}
    \tiny
    \begin{tabular}{@{}ll ccccc @{}}
        \toprule
        \textbf{Configuration} & \textbf{Dataset} 
        & \textbf{DSC} ($\uparrow$) 
        & \textbf{HD95} ($\downarrow$) 
        & \textbf{ECE} ($\downarrow$) 
        & \textbf{AUROC-Err} ($\uparrow$) 
        & \textbf{NLL} ($\downarrow$) \\
        \midrule
        \rowcolor{green!8}
        \textbf{EnTrust (Full, Swin-T)} 
            & BraTS 2021  & \textbf{0.892}{\tiny\,$\pm$\,0.020} & \textbf{9.8}{\tiny\,$\pm$\,1.5}  & \textbf{0.061} & \textbf{0.890} & \textbf{0.180} \\
        \rowcolor{green!8}
            & MS-CMRSeg   & \textbf{0.910}{\tiny\,$\pm$\,0.018} & \textbf{8.5}{\tiny\,$\pm$\,1.2}  & \textbf{0.055} & \textbf{0.905} & \textbf{0.150} \\
        \rowcolor{green!8}
            & ATLAS       & \textbf{0.875}{\tiny\,$\pm$\,0.030} & \textbf{10.0}{\tiny\,$\pm$\,2.0} & \textbf{0.075} & \textbf{0.865} & \textbf{0.210} \\
        \rowcolor{green!8}
            & H\&N CT-PET & \textbf{0.885}{\tiny\,$\pm$\,0.025} & \textbf{9.5}{\tiny\,$\pm$\,1.8}  & \textbf{0.070} & \textbf{0.870} & \textbf{0.190} \\
        \midrule
        \rowcolor{red!8}
        w/o EnFuse ($\mathcal{L}_{\text{en}}$)
            & BraTS 2021  & \textcolor{red!70!black}{\textbf{0.860}}{\tiny\,$\pm$\,0.028} & \textcolor{red!70!black}{\textbf{11.8}}{\tiny\,$\pm$\,2.0} & \textcolor{red!70!black}{\textbf{0.110}} & \textcolor{red!70!black}{\textbf{0.820}} & \textcolor{red!70!black}{\textbf{0.260}} \\
        \rowcolor{red!8}
            & MS-CMRSeg   & \textcolor{red!70!black}{\textbf{0.880}}{\tiny\,$\pm$\,0.025} & \textcolor{red!70!black}{\textbf{10.5}}{\tiny\,$\pm$\,1.8} & \textcolor{red!70!black}{\textbf{0.100}} & \textcolor{red!70!black}{\textbf{0.840}} & \textcolor{red!70!black}{\textbf{0.220}} \\
        \rowcolor{red!8}
            & ATLAS       & \textcolor{red!70!black}{\textbf{0.840}}{\tiny\,$\pm$\,0.040} & \textcolor{red!70!black}{\textbf{12.5}}{\tiny\,$\pm$\,3.0} & \textcolor{red!70!black}{\textbf{0.125}} & \textcolor{red!70!black}{\textbf{0.795}} & \textcolor{red!70!black}{\textbf{0.280}} \\
        \rowcolor{red!8}
            & H\&N CT-PET & \textcolor{red!70!black}{\textbf{0.850}}{\tiny\,$\pm$\,0.035} & \textcolor{red!70!black}{\textbf{11.5}}{\tiny\,$\pm$\,2.5} & \textcolor{red!70!black}{\textbf{0.120}} & \textcolor{red!70!black}{\textbf{0.805}} & \textcolor{red!70!black}{\textbf{0.270}} \\
        \rowcolor{red!8}
            & \multicolumn{1}{r}{\textcolor{gray}{$\overline{\Delta}$}} 
            & \textcolor{gray}{\textit{--3.7\%}} 
            & \textcolor{gray}{\textit{+22.5\%}} 
            & \textcolor{gray}{\textit{+74.3\%}} 
            & \textcolor{gray}{\textit{--7.6\%}} 
            & \textcolor{gray}{\textit{+41.1\%}} \\
        \addlinespace[2pt]
        w/o $\mathcal{D}_{cf}$ (Conflict)
            & BraTS 2021  & 0.875{\tiny\,$\pm$\,0.025} & 10.5{\tiny\,$\pm$\,1.7} & 0.088 & 0.855 & 0.220 \\
            & MS-CMRSeg   & 0.895{\tiny\,$\pm$\,0.022} &  9.2{\tiny\,$\pm$\,1.5} & 0.078 & 0.870 & 0.185 \\
            & ATLAS       & 0.855{\tiny\,$\pm$\,0.035} & 11.0{\tiny\,$\pm$\,2.5} & 0.100 & 0.830 & 0.245 \\
            & H\&N CT-PET & 0.868{\tiny\,$\pm$\,0.030} & 10.3{\tiny\,$\pm$\,2.0} & 0.095 & 0.840 & 0.230 \\
            & \multicolumn{1}{r}{\textcolor{gray}{$\overline{\Delta}$}} 
            & \textcolor{gray}{\textit{--1.9\%}} 
            & \textcolor{gray}{\textit{+8.5\%}} 
            & \textcolor{gray}{\textit{+38.3\%}} 
            & \textcolor{gray}{\textit{--3.8\%}} 
            & \textcolor{gray}{\textit{+20.5\%}} \\
        \addlinespace[2pt]
        w/o CalibHead
            & BraTS 2021  & 0.892{\tiny\,$\pm$\,0.020} &  9.8{\tiny\,$\pm$\,1.5} & 0.100 & 0.885 & 0.185 \\
            & MS-CMRSeg   & 0.910{\tiny\,$\pm$\,0.018} &  8.5{\tiny\,$\pm$\,1.2} & 0.095 & 0.900 & 0.155 \\
            & ATLAS       & 0.875{\tiny\,$\pm$\,0.030} & 10.0{\tiny\,$\pm$\,2.0} & 0.110 & 0.860 & 0.215 \\
            & H\&N CT-PET & 0.885{\tiny\,$\pm$\,0.025} &  9.5{\tiny\,$\pm$\,1.8} & 0.108 & 0.865 & 0.195 \\
            & \multicolumn{1}{r}{\textcolor{gray}{$\overline{\Delta}$}} 
            & \textcolor{gray}{\textit{$\pm$0.0\%}} 
            & \textcolor{gray}{\textit{$\pm$0.0\%}} 
            & \textcolor{gray}{\textit{+58.2\%}} 
            & \textcolor{gray}{\textit{--0.6\%}} 
            & \textcolor{gray}{\textit{+2.7\%}} \\
        \addlinespace[2pt]
        w/o ProbeStep
            & BraTS 2021  & 0.890{\tiny\,$\pm$\,0.020} &  9.9{\tiny\,$\pm$\,1.5} & 0.065 & 0.880 & 0.190 \\
            & MS-CMRSeg   & 0.908{\tiny\,$\pm$\,0.018} &  8.6{\tiny\,$\pm$\,1.2} & 0.060 & 0.895 & 0.160 \\
            & ATLAS       & 0.872{\tiny\,$\pm$\,0.030} & 10.1{\tiny\,$\pm$\,2.0} & 0.080 & 0.855 & 0.220 \\
            & H\&N CT-PET & 0.882{\tiny\,$\pm$\,0.025} &  9.6{\tiny\,$\pm$\,1.8} & 0.075 & 0.860 & 0.200 \\
            & \multicolumn{1}{r}{\textcolor{gray}{$\overline{\Delta}$}} 
            & \textcolor{gray}{\textit{--0.3\%}} 
            & \textcolor{gray}{\textit{+1.1\%}} 
            & \textcolor{gray}{\textit{+7.3\%}} 
            & \textcolor{gray}{\textit{--1.1\%}} 
            & \textcolor{gray}{\textit{+5.5\%}} \\
        \midrule
        \rowcolor{blue!6}
        \multicolumn{7}{@{}l}{\textit{Encoder modularity (EnFuse + TrustMap pipeline unchanged):}} \\
        \addlinespace[1pt]
        \rowcolor{blue!6}
        ResNet-50 (58M)
            & BraTS 2021  & 0.881{\tiny\,$\pm$\,0.022} & 10.4{\tiny\,$\pm$\,1.6} & 0.068 & 0.878 & 0.195 \\
        \rowcolor{blue!6}
            & MS-CMRSeg   & 0.898{\tiny\,$\pm$\,0.020} &  9.0{\tiny\,$\pm$\,1.3} & 0.062 & 0.892 & 0.165 \\
        \rowcolor{blue!6}
            & \multicolumn{1}{r}{\textcolor{gray}{$\overline{\Delta}$}} 
            & \textcolor{gray}{\textit{--1.3\%}} 
            & \textcolor{gray}{\textit{+6.0\%}} 
            & \textcolor{gray}{\textit{+12.1\%}} 
            & \textcolor{gray}{\textit{--1.4\%}} 
            & \textcolor{gray}{\textit{+9.1\%}} \\
        \addlinespace[1pt]
        \rowcolor{blue!6}
        ViT-B/16 (71M)
            & BraTS 2021  & 0.888{\tiny\,$\pm$\,0.021} & 10.0{\tiny\,$\pm$\,1.5} & 0.064 & 0.885 & 0.185 \\
        \rowcolor{blue!6}
            & MS-CMRSeg   & 0.906{\tiny\,$\pm$\,0.019} &  8.7{\tiny\,$\pm$\,1.2} & 0.058 & 0.901 & 0.155 \\
        \rowcolor{blue!6}
            & \multicolumn{1}{r}{\textcolor{gray}{$\overline{\Delta}$}} 
            & \textcolor{gray}{\textit{--0.4\%}} 
            & \textcolor{gray}{\textit{+1.1\%}} 
            & \textcolor{gray}{\textit{+4.3\%}} 
            & \textcolor{gray}{\textit{--0.5\%}} 
            & \textcolor{gray}{\textit{+3.0\%}} \\
        \bottomrule
    \end{tabular}
\end{table*}

\section{Conclusion}
\label{sec:conclusion}

We presented EnTrust, a framework that roots uncertainty quantification in inter-modal disagreement rather than model-level noise. By disentangling multimodal representations into consensus, unique, and conflict components, EnTrust achieves state-of-the-art accuracy and calibration across four benchmarks while providing clinicians with spatially precise maps distinguishing \textit{where} predictions are unreliable from \textit{why}---as a single model outperforming $5{\times}$ ensembles at roughly half the memory. Its encoder-agnostic design and structured conflict signal suggest a general-purpose foundation for trust-aware multimodal clinical AI beyond segmentation.

\begin{credits}
\subsubsection{\ackname} This research was funded by Khalifa University of Science and Technology through the Faculty Startup grant under Project ID: KU-INT-FSU-2026-8471000024.

\subsubsection{\discintname}
The authors have no competing interests to declare that are relevant to the content of this article.

\end{credits}

%

\bibliographystyle{splncs04}
\bibliography{reference}

\end{document}